\documentclass{article}
\usepackage[utf8]{inputenc}
\usepackage{authblk}

\usepackage{natbib}
\usepackage{hyperref}
\usepackage{mathrsfs}
\usepackage{amsmath}
\usepackage{amsfonts}
\usepackage{algorithm}
\usepackage{algpseudocode}
\usepackage{filecontents}
\usepackage{url}
\usepackage{subfig}
\usepackage{caption}
\usepackage{tikz,pgfplots,pgfplotstable,tkz-graph,tikz-dependency}

\newcommand{\ie}{i.e., }

\newcommand{\EQ}[1]{Equation~\ref{#1}}
\newcommand{\FIG}[1]{Figure~\ref{#1}}
\newcommand{\ALG}[1]{Algorithm~\ref{#1}}
\newcommand{\TABLE}[1]{Table~\ref{#1}}

\newcommand{\norm}[1]{\left\lVert#1\right\rVert}

\newcommand{\E}[1]{\mbox{\bf E}[#1]}
\newcommand{\rv}[1]{\mathbf{#1}}
\newcommand{\expnum}[2]{{#1}\mathrm{e}{#2}}

\title{Shifted Randomized Singular Value Decomposition}
\author{Ali Basirat\thanks{ali.basirat@lingfil.uu.se}}
\affil{Department of Linguistics and Philology\\ Uppsala University}
\date{August 2019}

\begin{document}

\maketitle

\begin{abstract}
  We extend the randomized singular value decomposition (SVD) algorithm \citep{Halko2011finding} to estimate the SVD of a shifted data matrix without explicitly constructing the matrix in the memory.
  With no loss in the accuracy of the original algorithm, the extended algorithm provides for a more efficient way of matrix factorization. 
  The algorithm facilitates the low-rank approximation and principal component analysis (PCA) of off-center data matrices.
  When applied to different types of data matrices, our experimental results confirm the advantages of the extensions made to the original algorithm.
\end{abstract}

\section{Introduction}
The singular value decomposition (SVD) is one of the most used matrix decompositions in many areas of science.
Among the typical applications of SVD are the low-rank matrix approximation and principal component analysis (PCA) of data matrices \citep{jolliffe2002principal}.
Using SVD to accurately estimate a low-rank factorization or the principal components of a data matrix, a mean-centering step should be carried out before performing SVD on the matrix.       
Despite its simplicity, the mean-centering can be very costly if the data matrix is large and sparse.
This cost is because the mean subtraction of a sparse matrix turns it to a dense matrix which requires a considerable amount of memory and CPU time to be analyzed.        
This motivates us to extend the randomized SVD algorithm introduced by \citep{Halko2011finding} to estimate the singular value decomposition of a mean-centered matrix without explicitly forming the matrix in the memory.

More generally, we introduce a shifted randomized SVD algorithm that provides for the SVD estimation of a data matrix shifted by any vector in the eigenspace of its column vectors.                                
The proposed algorithm facilitates the low-rank matrix approximation and the principal component analysis of a data matrix through merging the mean-centering and the SVD steps.
The mean-centering is crucial to obtain the minimum PCA reconstruction error through a deterministic SVD.
We experimentally show that it plays an essential role in case of using the randomized SVD algorithm too.               
Our experiments with different types of data matrices show that the extended algorithm performs better than the original algorithm when both are applied to a center-off data matrix.

In the followings, we briefly introduce the principal component analysis and its connection with the singular value decomposition. 
Then, we introduce the shifted randomized SVD algorithm and provide an analysis of its performance. 
Finally, we report our experimental results obtained from the principal component analysis of different types of data matrices using the extended and the original randomized SVD algorithms.

\section{Principal Component Analysis}
\label{sec:pca}

Principal component analysis (PCA) is a method to study the variance of a random vector.                               
PCA projects a random vector to the eigenspace of its covariance matrix.
Let $\rv{x}$ be an $m$-dimensional random vector with the mean vector $\rv{0}$.
PCA projects $\rv{x}$ to a latent random vector $\rv{y}$ as below:
\begin{equation}
  \rv{y}=A^T \rv{x}
  \label{eq:pca_projection}
\end{equation}
where the square matrix $A$ is composed of the eigenvectors of the covariance matrix $\Sigma_\rv{x}$.
The elements of $\rv{y}$ are called the \emph{principal components} of $\rv{x}$.
In many use cases of PCA, $A$ contains a subset of the eigenvectors of $\Sigma_\rv{x}$.
The minimum PCA reconstruction error is obtained from the eigenvectors corresponding to the top eigenvalues of the covariance matrix \citep{jolliffe2002principal}.

The matrix of eigenvectors $A$ in \EQ{eq:pca_projection} can be efficiently estimated from the singular value decomposition (SVD) of a sample matrix $X$. 
To this end, the sample matrix should be first centered around its mean vector:
\begin{equation}
  \bar{X} = X - \mu_\rv{x}\rv{1}^T
  \label{eq:sample_mean_centring}
\end{equation}
The fact that the matrix of left singular vectors of $\bar{X}$ is equal to the eigenvectors of the covariance matrix of $X$, the PCA projection of $X$ is:
\begin{equation}
  Y =  U^T\bar{X} = SV^T 
  \label{eq:pca_svd}
\end{equation}
where $\bar{X}=USV^T$. 

The mean-centering step in \EQ{eq:sample_mean_centring} can be very costly if $X$ is a large sparse matrix and $\mu_\rv{x}\ne\rv{0}$. 
In this case, $\bar{X}$ is a dense matrix that requires a vast amount of memory and cannot be processed in a reasonable time.
In the next section, we introduce a randomized SVD algorithm to estimate the  SVD of $\bar{X}$ without explicitly performing the mean subtraction step.

\section{Shifted Singular Value Decomposition}
\label{subsec:svd}
Let $X$ be an $m \times n$ $(m \le n)$ matrix and $\mu$ be an $m$ dimensional vector in the space of the column vectors of $X$.
\ALG{alg:rsvd}, extends the randomized matrix factorization method introduced by \citet{Halko2011finding} to return a rank-$k$ approximation of the singular value decomposition of the matrix $\bar{X} = X - \mu\rv{1}^T$ without explicitly forming $\bar{X}$.
The differences between the extended algorithm and the original one are in lines \ref{alg:rsvd:qrupdate}, \ref{alg:rsvd:for1}, \ref{alg:rsvd:for2}, and \ref{alg:rsvd:projection}.
In the followings, we explain the entire algorithm with a more in-depth focus on the modified parts.

The \textsc{Shifted-Randomized-SVD} algorithm works in three major steps:
\begin{enumerate}
  \item Estimate a basis matrix for $\bar{X}$
  \item Project $\bar{X}$ to the space of the basis matrix
  \item Estimate the SVD factors of $\bar{X}$ from its projection
\end{enumerate}
In the first step (lines \ref{alg:rsvd:omega} to \ref{alg:rsvd:endfor}), a rank $K$ basis matrix $Q_1$ ($k<K\ll m$) that spans the column vectors of $\bar{X}$ is computed.
In Line~\ref{alg:rsvd:omega}, a random matrix is drawn from the standard Gaussian distribution.
This matrix is then used in Line~\ref{alg:rsvd:smapling} to form the sample matrix $X_1$ whose columns are independent random points in the range of $X$.
This sample matrix is used to estimate a basis matrix for $\bar{X}$ in two steps.
In Line~\ref{alg:rsvd:qr}, a basis matrix $Q_1$ is computed through QR-factorization of $X_1$.
Since $X_1$ is sampled from $X$, the basis matrix is considered as an approximation of the basis of $X$.
Then in Line~\ref{alg:rsvd:qrupdate}, the basis of $\bar{X}$ is estimated from the $Q_1$ by the QR-update algorithm proposed by \citet[p.~607]{Golub:MC}.
For a given QR factorization such as $Q_1R_1=X_1$ and two vectors $u$, and $v$, the QR-update algorithm computes the QR-factorization in \EQ{eq:qrupdate} by updating the already available factors $Q_1$ and $R_1$.
\begin{equation}
  QR=X_1+uv^T
  \label{eq:qrupdate}
\end{equation}
The computational complexity of the QR-update of the $m\times K$ matrix $X_1$ is $O(m^2)$.\footnote{The computational complexity of the QR-update of an $m \times n$ matrix in $O(N^2)$ where $N=\mbox{max}(m,n)$.}
Replacing $u$ with $-\mu$ and $v$ with $\rv{1}$, the QR-update in Line~\ref{alg:rsvd:qrupdate} returns the basis matrix $Q$ that spans the range of the matrix
\begin{equation}
  \bar{X} = X - \mu\rv{1}^T
\end{equation}
In other words, $\bar{X}$ can be approximated from $Q$:
\begin{equation}
  \bar{X} \approx QQ^T\bar{X}
  \label{eq:X_approximation}
\end{equation}
Note that the basis matrix of $\bar{X}$ is computed without explicitly constructing the matrix $\bar{X}$ itself.
The \texttt{if} statement in Line~\ref{alg:rsvd:qr} controls the useless performance of the QR-update step with the null vector. 

The \texttt{for} loop starting at Line~\ref{alg:rsvd:for}, estimates a basis matrix for $B=(\bar{X}\bar{X}^T)^q\bar{X}$ using the basis of $\bar{X}$, $Q$.
The matrix $B$ with a positive integer power has the same singular vectors as $\bar{X}$, but with a sharper decay in its singular values since $s_j(B)=s_j(\bar{X})^{2q+1}$, where $s_j(.)$ returns the $j$th singular vectors of its input matrix.
The sharp decay in singular values improves the accuracy of the randomized SVD when the singular values of $\bar{X}$ decay slowly.
This effect is because the reconstruction error of the randomized SVD is directly proportional to the first top unused singular vector of $\bar{X}$ (see \EQ{eq:err}).

The basis of $(\bar{X}\bar{X}^T)^q\bar{X}$ is computed via alternative applications of matrix product on $\bar{X}^T$ and $\bar{X}$.
For $q=1$, in Line~\ref{alg:rsvd:for1}, a basis matrix of $\bar{X}^T\bar{X}$ is estimated through QR-factorization of $\bar{X}^TQ$.
To avoid forming $\bar{X}$ explicitly, instead of direct multiplication $\bar{X}^TQ$, we use the distributive property of multiplication over addition:
\begin{equation}
  \bar{X}^TQ = (X - \mu\rv{1}^T)^TQ = X^TQ - \rv{1}(\mu^T Q)
  \label{eq:alt_mult_1}
\end{equation}
where $\rv{1}$ is a vector of ones. 
The product $\rv{1}(\mu^T Q)$ can be efficiently computed in $O(nK)$ memory space if a higher priority is given to the parentheses. 
In Line~\ref{alg:rsvd:for2}, a basis matrix of $\bar{X}\bar{X}^T\bar{X}$ is estimated through QR-factorization of $\bar{X}Q'$ where $Q'$ is a basis matrix of $\bar{X}^TQ$. 
Similar to \EQ{eq:alt_mult_1}, the product $\bar{X}Q'$ is computed as:
\begin{equation}
  \bar{X}Q' = (X - \mu\rv{1}^T)Q' = XQ' - \mu(\rv{1}^TQ')
  \label{eq:alt_mult_1}
\end{equation}
with the same amount of memory space, $O(nK)$. 
The matrix multiplication loop iterates $q$ times.
At this stage, we have the basis matrix $Q$ that approximates a basis for $\bar{X}$. 

In the second major step of the algorithm, the matrix $\bar{X}$ is projected to the space spanned by $Q$:
\begin{equation}
  Y = Q^T\bar{X}
  \label{eq:reprojection}
\end{equation}
This step in done in Line~\ref{alg:rsvd:projection} using the same trick as in \EQ{eq:alt_mult_1}:
\begin{equation}
  Y = Q^T(X - \mu\rv{1}^T) = Q^TX - (Q^T\mu)\rv{1}^T
  \label{eq:reprojection_expanded}
\end{equation}

Finally, in the third step, the SVD factors of $\bar{X}$ are estimated from the $K \times n$ matrix $Y$ in two steps.  
First, a rank-$k$ approximation of $Y$ is computed using a standard method of singular value decomposition, \ie $Y=U_{1}\Sigma V^T$ (Line~\ref{alg:rsvd:svd}). 
Then, the left singular vectors are updated by $U = QU_{1}$ resulting in $U\Sigma V^T=QY$ (Line~\ref{alg:rsvd:qu}).  
Replacing $Y$ with $Q^T\bar{X}$ and using \EQ{eq:X_approximation} ($\bar{X} \approx QQ^T\bar{X}$), we have the rank-$k$ approximation of $\bar{X}$:
\begin{equation}
  U\Sigma V^T\approx\bar{X}
  \label{eq:shifted_randomized_svd}
\end{equation}

\begin{algorithm}[H]
  \begin{algorithmic}[1]
    \Procedure{Shifted-Randomized-SVD}{$X,\mu,k,K,q$} 
      \State Draw an $n \times K$ standard Gaussian matrix $\Omega$ \label{alg:rsvd:omega}
      \State Form the sample matrix $X_1 \gets X\Omega$ \label{alg:rsvd:smapling} 
      \State Compute the QR factorization $X_1 = Q_1R_1$ \label{alg:rsvd:qr}
      \If{$\mu\ne\rv{0}$} \label{alg:rsvd:if}
        \State Compute $QR = Q_1R_1 - \mu\rv{1}^T$ using the QR-update algorithm \label{alg:rsvd:qrupdate}
      \EndIf
      \For{$i=1,2,\dots,q$} \label{alg:rsvd:for}
        \State Compute the QR-factorization $Q'R' = X^TQ -\rv{1}(\mu^TQ)$ \label{alg:rsvd:for1}
        \State Compute the QR-factorization $QR = XQ'-\mu(\rv{1}^TQ')$ \label{alg:rsvd:for2}
      \EndFor \label{alg:rsvd:endfor}
      \State Form $Y \gets Q^TX - (Q^T\mu)\rv{1}^T$   \label{alg:rsvd:projection}
      \State Compute the singular value decomposition of $Y = U_{1}\Sigma V^T$ \label{alg:rsvd:svd}
      \State $U \gets QU_{1}$         \label{alg:rsvd:qu}
      \State \Return $(U, \Sigma, V)$
    \EndProcedure
  \end{algorithmic}
  \caption{The rank-$k$ singular value decomposition of the $m\times n$ matrix $X-\mu\rv{1}^T=U\Sigma V^T$ with $(m \le n)$ using the sampling parameter $K$ $(k < K \ll m)$ and $q\in\{0,1,2,\dots\}$.}
\label{alg:rsvd}
\end{algorithm}

The shifting vector $\mu$ in \ALG{alg:rsvd} can be any vector in the space of the column vectors of $X$. 
If it is set to the null vector $\rv{0}$, then the algorithm reduces to the original randomized SVD algorithm of \citet{Halko2011finding}. 
If it is set to the mean vector of $X$, then the algorithm estimates the singular vectors of the mean-centered matrix $\bar{X}$.
In this case, the algorithm facilitates the principal component analysis of a data matrix $X$ through merging the centering step in \EQ{eq:sample_mean_centring} and the SVD step in \EQ{eq:pca_svd}. 

\section{Performance Analysis}
The \textsc{Shifted-Randomized-SVD} algorithm explained in the previous section approximates the SVD of a shifted data matrix $\bar{X}=X-\mu\rv{1}^T$ without explicitly constructing the matrix in the memory.
In this section, we study the performance of the algorithm based on the accuracy and the efficiency of the original randomized SVD algorithm of \citet{Halko2011finding}. 

To estimate the singular value decomposition of a shifted matrix $\bar{X}$ using the original randomized SVD algorithm, $\bar{X}$ should be explicitly formed and passed to the algorithm. 
Since \textsc{Shifted-Randomized-SVD} adds no extra randomness to the original algorithm, we have the same reconstruction error bound as if the original algorithm factorized the shifted matrix $\bar{X}$ \citep{Halko2011finding}:
\begin{equation}
  \E{\norm{\bar{X}-USV^T}}\le \left[1+4\sqrt{\frac{2m}{k-1}}\right]^\frac{1}{2q+1}\sigma_{k+1}
  \label{eq:err}
\end{equation}
where $\sigma_{k+1}$ is the $(k+1)$th singular value of the $m\times n$ matrix $\bar{X}$ with $m\le n$, $2 \le k \le \frac{m}{2}$ is the decomposition rank, and $q\in\mathbb{Z}^+$ is a power value as explained in \ALG{alg:rsvd}. 

In the followings, we study the computational complexity of the SVD factorization of $\bar{X}$ using the original randomized SVD algorithm and its extended version in \ALG{alg:rsvd}. 
For an $m\times n$ matrix $\bar{X}$, the computational complexity of the original randomized SVD algorithm is:
\begin{equation}
  O(\alpha k + (m+n)k^2)
  \label{eq:time_complexity}
\end{equation}
where $\alpha$ is the cost of the matrix-vector multiplication with the input matrix $\bar{X}$.
If $\bar{X}$ is a dense matrix then $\alpha = mn$ , and if $\bar{X}$ is a sparse matrix then $\alpha = T$, a small constant value. 

\ALG{alg:rsvd} adds a QR-update step (Line~\ref{alg:rsvd:qrupdate}) and three matrix-matrix multiplications (lines \ref{alg:rsvd:for1}, \ref{alg:rsvd:for2}, and \ref{alg:rsvd:projection}) to the original algorithm. 
The matrix multiplications do not affect the computational complexity of the original algorithm since their computational complexity is equal to the complexity of computing $Q^TX$ in the original algorithm.  
The QR-update step, running in $O(m^2)$, however, can affect the computational complexity of the algorithm. 

Assuming that $\mu\ne\rv{0}$, if both $X$ and $\bar{X}$ are dense matrices then both algorithms have the same computational complexity as:
\begin{equation}
  O(mnk + (m+n)k^2)
  \label{eq:time_complexity}
\end{equation}
This equality is because the computational complexity of the QR-update step, $O(m^2)$, is dominated by the complexity of the original algorithm (\ie $m^2 \le mn$ for every $m \le n$ where $m,n\in\mathbb{N}$). 
In addition, the construction of $\bar{X}$ to be used by the original algorithm takes $O(mn)$ time which is greater than or equal to the complexity of the QR-update step. 
Hence, the added operations do not affect the computational complexity of the original algorithm.
\citet{Halko2011finding} show that for a dense input matrix, the randomized SVD algorithm can be performed in $O(mn\log{k} + (m+n)k^2)$ if instead of the random normal matrix $\Omega$ in Line~\ref{alg:rsvd:omega}, a structured random matrix such as the sub-sampled random Fourier transform is used. 
This improvement can be considered for the \textsc{Shifted-Randomized-SVD} algorithm too. 

If the input matrix $X$ is sparse, then $\bar{X}$ is a dense matrix for every $\mu\ne\rv{0}$. 
In this case, the computational complexity of the \textsc{Shifted-Randomized-SVD} algorithm is:
\begin{equation}
  O(Tk + m^2 + (m+n)k^2)
\end{equation}
where the constant $T$ is the cost of multiplying a sparse matrix to a vector, and the parameter $m^2$ is related to the complexity of the QR-update step. 
On the other hand, since $\bar{X}$ is a dense matrix, the complexity of the original algorithm is $O(mnk + (m+n)k^2)$ which is higher than the complexity of the extended algorithm. 

In a special case where $X$ is a dense matrix and $\bar{X}$ is a sparse matrix, the original algorithm can factorize $\bar{X}$ in $O(Tk + (m+n)k^2)$, but \textsc{Shifted-Randomized-SVD} needs $O(mnk + (m+n)k^2)$ time.
In this case, if \ALG{alg:rsvd} is applied to $\bar{X}$ with $\mu=\rv{0}$, the factorization can be performed in the same processing time as the original algorithm. 
As a summary, we showed that the \textsc{Shifted-Randomized-SVD} algorithm as illustrated in \ALG{alg:rsvd} is as efficient as the randomized SVD algorithm proposed by \citet{Halko2011finding} if the input matrix is dense, and more efficient than it if the input matrix is sparse. 

\section{Experiments}

\pgfplotstableread[col sep=&,header=false]{
1 & 8.22704387130385
2 & 8.131393039362736
3 & 8.016755695600061
4 & 7.923705597236911
5 & 7.828101953480554
10 & 7.326654184002265
20 & 6.251097197412579
30 & 5.193074751939208
40 & 4.101132326695938
50 & 3.0243988000795885
60 & 2.2630846386145804
70 & 1.5815823996627905
80 & 0.9732444268685795
90 & 0.4409804224778575
}\centredReconstrucionErrors

\pgfplotstableread[col sep=&,header=false]{
1 & 14.358630406889038
2 & 9.636071404548272
3 & 9.082191191384167
4 & 8.850339522665823
5 & 9.577975663349719
10 & 7.5724349048787065
20 & 6.338675405968799
30 & 5.303208301235163
40 & 4.128050972158258
50 & 3.0389259384176377
60 & 2.2735689661599277
70 & 1.5897759227859083
80 & 0.9801874908151821
90 & 0.4453492483439178
}\uncentredReconstrucionErrors

\pgfplotstableread[col sep=&,header=false]{
1000 & 71.4485171766514
2000 & 73.49453380835662
3000 & 74.15975028355514
4000 & 74.85221486980839
5000 & 75.16547637478845
7000 & 75.54367178701746
10000 & 75.87600056334387
15000 & 76.20677027707752
20000 & 76.49794058022411
}\centredSampleSizeRE

\pgfplotstableread[col sep=&,header=false]{
1000 & 78.03850122793364
2000 & 80.57495504280583
3000 & 89.91490502714504
4000 & 81.9055564379236
5000 & 89.13459199773038
7000 & 84.58760839961361
10000 & 88.16082999248255
15000 & 89.90041660505608
20000 & 95.83943271534072
}\uncentredSampleSizeRE

\pgfplotstableread[col sep=&,header=false]{
Uniform & 70.9727066912633
Normal & 855.8096731941326
Zipf & 2134.25260619535
Poisson & 854.8843992368012
}\centredDistRE

\pgfplotstableread[col sep=&,header=false]{
Uniform & 88.78436797086918
Normal & 1039.3295354374552
Zipf & 2459.7144493445953
Poisson & 1045.3963120003946
}\uncentredDistRE

\pgfplotstableread[col sep=&,header=false]{
Uniform & 71.5149824196122
Normal & 863.2543530919766
Zipf & 1989.8301637177353
Poisson & 861.3224353737171
}\ImplicitCentredDistRE

\pgfplotstableread[col sep=&,header=false]{
Uniform & 71.48041234638316
Normal & 863.345200515253
Zipf & 1968.193909071475
Poisson & 860.9032427038136
}\ExplicitCentredDistRE

\pgfplotstableread[col sep=&,header=false]{
0 & 71.31809930585888
1 & 70.83076312467998
2 & 70.27761914740873
3 & 70.18890364592957
4 & 69.81530304604695
5 & 69.91298789391546
10 & 69.65911175471719
20 & 69.32992309328083
30 & 69.64508487832398
40 & 69.39384648077716
50 & 69.43326583466681
100 & 69.48461156650795
200 & 69.37689031371065
}\QUnifCentred

\pgfplotstableread[col sep=&,header=false]{
0 & 98.65144901586238
1 & 71.15026418515853
2 & 70.61986766065073
3 & 70.55413230089404
4 & 70.15943619392725
5 & 70.337018844345
10 & 70.14369280367373
20 & 69.71064800826655
30 & 70.13059675593622
40 & 69.86166061002311
50 & 69.876535100108
100 & 69.87456445172795
200 & 69.8096471733457
}\QUnifUncentred

\pgfplotstableread[col sep=&,header=true]{
  distribution & q & diff
Uniform & 0 & -27.333349710003503
Uniform & 1 & -0.319501060478558
Uniform & 2 & -0.3422485132419979
Uniform & 3 & -0.36522865496446855
Uniform & 4 & -0.34413314788029936
Uniform & 5 & -0.42403095042953964
Uniform & 10 & -0.4845810489565423
Uniform & 20 & -0.3807249149857199
Uniform & 30 & -0.4855118776122396
Uniform & 40 & -0.46781412924595145
Uniform & 50 & -0.44326926544118805
Uniform & 100 & -0.3899528852199978
Uniform & 200 & -0.43275685963504884
}\QCentMinusUncentUniform
\pgfplotstableread[col sep=&,header=true]{
  distribution & q & diff
Normal & 0 & -102.63177661229679
Normal & 1 & -3.147389474049419
Normal & 2 & -3.932107702159783
Normal & 3 & -3.9757769624708317
Normal & 4 & -4.642661505139927
Normal & 5 & -4.806349322319534
Normal & 10 & -5.039278153138866
Normal & 20 & -6.00605867759873
Normal & 30 & -4.8565526848095715
Normal & 40 & -5.062091226355506
Normal & 50 & -4.354403341247462
Normal & 100 & -5.062091226355506
Normal & 200 & -4.877435163736436
}\QCentMinusUncentNormal
\pgfplotstableread[col sep=&,header=true]{
  distribution & q & diff
Zipf & 0 & -239.42544596388234
Zipf & 1 & -551.8677405074009
Zipf & 2 & -37.08394303035129
Zipf & 3 & -116.78370350398632
Zipf & 4 & -546.3583027964987
Zipf & 5 & -111.5165017019292
Zipf & 10 & -42.77065404711402
Zipf & 20 & -48.62856840818745
Zipf & 30 & -247.01753774032568
Zipf & 40 & -143.60674276270515
Zipf & 50 & -56.4398721125599
Zipf & 100 & -64.35828332054189
Zipf & 200 & -63.54045191399541
}\QCentMinusUncentZipf
\pgfplotstableread[col sep=&,header=true]{
  distribution & q & diff
Poisson & 0 & -175.24587897190952
Poisson & 1 & -2.288230291040122
Poisson & 2 & -3.008317607061713
Poisson & 3 & -3.6423137715332814
Poisson & 4 & -4.528759577336814
Poisson & 5 & -4.440253487754035
Poisson & 10 & -5.373274059620712
Poisson & 20 & -5.322577917957915
Poisson & 30 & -5.218599994522947
Poisson & 40 & -5.159822099602934
Poisson & 50 & -5.17988535199413
Poisson & 100 & -4.929459207756167
Poisson & 200 & -5.322591505297851
}\QCentMinusUncentPoisson

We experimentally study the difference between performing PCA with the randomized SVD algorithm (RSVD) proposed by \citet{Halko2011finding} and its extended version in \ALG{alg:rsvd} (S-RSVD). 
The fact that a minimum PCA reconstruction error is obtained from the deterministic SVD of a mean-centered data matrix, the performance of S-RSVD on an off-center data matrix with its mean vector as the shifting vector is expected to be more accurate than the performance of RSVD on the same data matrix. 
Since the two algorithms are randomized in nature, it is important to test if this expectation is valid for different types of data matrices.

We experimentally compare the two algorithms based on the mean of the squared $L_2$ norm of PCA reconstruction error (MSE). 
The same parameters $K=2k$ and $q=0$ are used for both S-RSVD and RSVD unless it is clearly mentioned. 
The shifting vector $\mu$ for S-RSVD is set to the mean vector of data matrices. 
The experiments are carried out on different types of data matrices including randomly generated data, a set of images, and word co-occurrence matrices.
The characteristics of the data matrices are illustrated in the corresponding sections. 

\subsection{Random Data}

In this section, we examine how the two SVD algorithms are affected by the parameters such as the number of principal components, the size and the distribution of a data matrix, and the power iteration scheme. 
Our experiments are based on two comparison metrics, 1) an MSE value obtained from a fixed number of principal components, and 2) the sum of MSE values obtained from different number of principal components ranging from $1$ to $100$. 

\begin{figure}
  \centering
  \subfloat[][] {
    \begin{tikzpicture}
      \begin{axis}[
          width=0.5\textwidth,
          height=3cm,
          tick label style={font=\tiny},
          legend style={font=\tiny, draw=none, fill=none},
        ]
        \addplot [mark=none] table[x index=0, y index=1] \centredReconstrucionErrors ; \addlegendentry{S-RSVD} ;
        \addplot [mark=none, dashed] table[x index=0, y index=1] \uncentredReconstrucionErrors ;\addlegendentry{RSVD} ;
      \end{axis}
    \end{tikzpicture}
    \label{fig:re}
  }
  \subfloat[][] {
    \begin{tikzpicture}
      \begin{axis}[
          scaled ticks=base 10:3,
          scaled y ticks = false,
          scaled x ticks = false,
          legend style={font=\tiny, draw=none, fill=none},
          height=3cm,
          width=0.5\textwidth,
          tick label style={font=\tiny},
        ]
        \addplot [mark=none] table[x index=0, y index=1] \centredSampleSizeRE ; \addlegendentry{S-RSVD} ;
        \addplot [mark=none, dashed] table[x index=0, y index=1] \uncentredSampleSizeRE ;\addlegendentry{RSVD} ;
      \end{axis}
    \end{tikzpicture}
    \label{fig:sample_size}
  }

  \subfloat[][] {
    \begin{tikzpicture}
      \begin{axis}[
         ybar,
         width=0.5\textwidth,
         height=3cm,
         tick label style={font=\tiny},
         symbolic x coords={Uniform, Normal, Zipf, Poisson},
         y tick label style={},
         xtick=data,
         xtick pos=left,
         ymode=log,
         legend style={at={(0.02,0.98)},anchor=north west, font=\tiny, fill=none, draw=none},
        ]
        \addplot [pattern=north west lines] table[x index=0, y index=1] \centredDistRE ; \addlegendentry{S-RSVD} ;
        \addplot [pattern=north east lines] table[x index=0, y index=1] \uncentredDistRE ;\addlegendentry{RSVD} ;
      \end{axis}
    \end{tikzpicture}
    \label{fig:data_dist}
  }
  \subfloat[][] {
    \begin{tikzpicture}
      \begin{axis}[
         ybar,
         width=0.5\textwidth,
         height=3cm,
         tick label style={font=\tiny},
         symbolic x coords={Uniform, Normal, Zipf, Poisson},
         y tick label style={},
         xtick=data,
         xtick pos=left,
         ymode=log,
         legend style={at={(0.02,0.98)},anchor=north west, font=\tiny, fill=none, draw=none},
        ]
        \addplot [pattern=north west lines] table[x index=0, y index=1] \ImplicitCentredDistRE ; \addlegendentry{S-RSVD} ;
        \addplot [pattern=north east lines] table[x index=0, y index=1] \ExplicitCentredDistRE ;\addlegendentry{RSVD} ;
      \end{axis}
    \end{tikzpicture}
    \label{fig:imp_exp}
  }

  \subfloat[][] {
    \begin{tikzpicture}
      \begin{axis}[
          width=0.5\textwidth,
          height=3cm,
          tick label style={font=\tiny},
          legend style={font=\tiny, draw=none, fill=none},
        ]
        \addplot [mark=none] table[x index=0, y index=1] \QUnifCentred; \addlegendentry{S-RSVD} ;
        \addplot [mark=none, dashed] table[x index=0, y index=1] \QUnifUncentred ;\addlegendentry{RSVD} ;
      \end{axis}
    \end{tikzpicture}
    \label{fig:q}
  }
  \subfloat[][] {
    \begin{tikzpicture}
      \begin{axis}[
          width=0.5\textwidth,
          height=3cm,
          tick label style={font=\tiny},
          legend style={font=\tiny, draw=none, fill=none},
        ]
        \addplot [mark=none, densely dotted] table[x=q, y=diff] \QCentMinusUncentUniform ; \addlegendentry{Uniform} ;
        \addplot [mark=none, dashed] table[x=q, y=diff] \QCentMinusUncentNormal ; \addlegendentry{Normal} ;
        \addplot [mark=none, solid] table[x=q, y=diff] \QCentMinusUncentZipf ; \addlegendentry{Zipf} ;
        \addplot [mark=none, dotted] table[x=q, y=diff] \QCentMinusUncentPoisson ; \addlegendentry{Poisson} ;
      \end{axis}
    \end{tikzpicture}
    \label{fig:q_dist}
  }

  \caption{A comparison between S-RSVD and RSVD based on (a) number of principal components, (b) sample size, (c) data distribution, (d) explicit versus implicit mean-centering, (e) the power value $q$, and (f) the difference of their accuracies with respect to the power value $q$ and the data distribution.}
  \label{fig:random_data}
\end{figure}
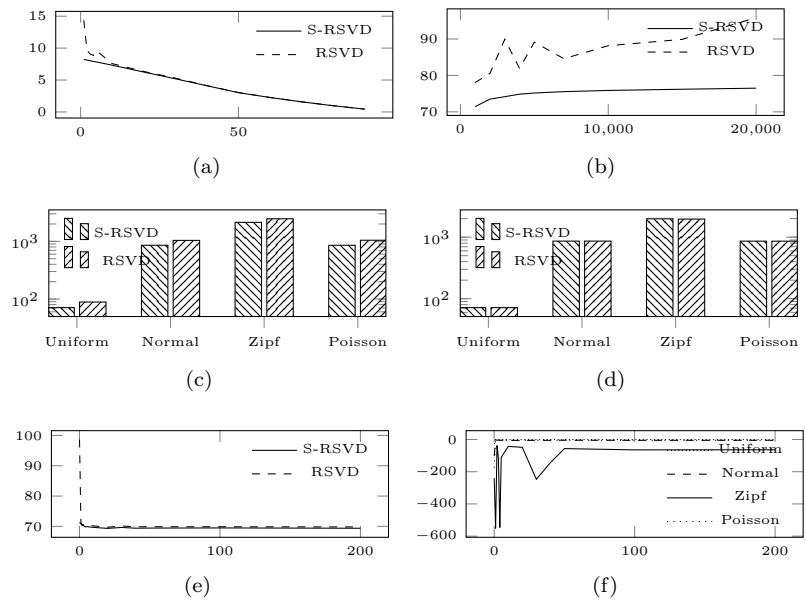

\FIG{fig:re} represents the effect of the number of principal components on the MSE values obtained from a $100 \times 1000$ matrix sampled from a $100$-dimensional random vector uniformly distributed in the range $\rv{0},\rv{1}$. 
The results show that mean-centering leads to substantial reduction to the reconstruction error when the number of principal components is small. 
This observation is in line with the fact that the contribution of the mean-centering is mostly to the accuracy of top principal components. 

The effect of the sample size on the two factorization algorithms is represented in \FIG{fig:sample_size} in which the X-axis is the sample size and the Y-axis is the sum of the MSE values obtained from different number of principal components ranging from $1$ to $100$. 
The data matrices are generated by a $100$-dimensional uniform random vector in the range $\rv{0},\rv{1}$. 
The results show that S-RSVD is more accurate and stable than RSVD.
Despite the fact that both algorithms are randomized, the stability of S-RSVD is less sensitive to the sample size. 

\FIG{fig:data_dist} compares the performance of the two algorithms with respect to the data distribution. 
The Y-axis is the sum of MSE values.
Regardless of the data distribution, S-RSVD is more accurate than RSVD. 
This observation is in line with the fact that PCA does not make any assumption about the data distribution. 

To examine whether both algorithms are equally accurate for the factorization of a mean-centered data matrix $\bar{X}$, a comparison between the two algorithms is provided in \FIG{fig:imp_exp}. 
The Y-axis is the sum of the MSE values.
In the experiments with RSVD, the matrix $\bar{X}$ is explicitly constructed and factorized, but in the S-RSVD experiments the singular factors of $\bar{X}$ are implicitly estimated from $X$. 
The results show that S-RSVD is as accurate as RSVD applied to an already centered data matrix $\bar{X}$. 
This observation supports \EQ{eq:shifted_randomized_svd}. 

An important parameter of the randomized SVD algorithm is the power value $q$ that has a positive effect on the accuracy of the algorithm, .
\FIG{fig:q} shows the MSE values obtained from each of the factorization algorithms with different values of $q$. 
The data matrix in this experiment is sampled from a $100$-dimensional uniform distribution. 
The Y-axis is the sum of MSE values and the X-axis represents $q$. 
The importance of mean-centering is clear when the value of $q$ is small. 
The accuracy of RSVD is significantly improved as the values of $q$ increases, while the accuracy of S-RSVD is only slightly improved. 
This observation on a set of uniformly distributed vectors suggests that RSVD with a positive value of $q$ ($1$ or $2$ as suggested by \citet{Halko2011finding}) can be as accurate as S-RSVD. 

To test whether RSVD with a large value of $q$ can be as accurate as S-RSVD, we run the same experiment as above but on data with different distributions.
\FIG{fig:q_dist} shows the difference between the sum of MSE values obtained from each of the algorithms (\ie Y-axis is $\mbox{MSE-SUM}(\mbox{S-RSVD}) - \mbox{MSE-SUM}(\mbox{S-RSVD})$) with respect to the parameter $q$. 
Being all the results negative means that S-RSVD is more accurate than RSVD. 
Except for the data with Zipfian distribution, the difference between the accuracy of the two algorithms approaches to zero as the value of $q$ increases.
The Zipfian graph fluctuates widely for small values of $q$, but it becomes flat as $q$ becomes larger.
In the best case, the difference between the two algorithms on Zipfian data is $-64$ at $q=200$.  
This indicates that the power iteration scheme cannot fully recover the reconstruction loss of an off-center data matrix, but depending on the data distribution, it can be helpful. 
The power iteration in \ALG{alg:rsvd} is a computationally heavy step which can negatively affect the efficiency of the algorithm when the value of $q$ is large. 

\subsection{Image Data}
\label{sec:img}
In this section, we experiment with handwritten digits and facial image matrices.
The handwritten digits are a copy of the test set of the UCI ML hand-written digits datasets consisting of $1979$ images of size $8\times8$.\footnote{\url{https://archive.ics.uci.edu/ml/datasets/Optical+Recognition+of+Handwritten+Digits}}
We vectorize individual image matrices and stack them into a single $64\times1979$ data matrix. 
The facial images consisting of $13233$ images each of size $250\times250$ are downloaded from Labeled Faces in the Wild (LFW).\footnote{\url{http://vis-www.cs.umass.edu/lfw/lfw.tgz}}
The facial image matrix after vectorizing and stacking all of the images matrices is a $62500\times13233$ matrix. 
 
The left side of \TABLE{MSE_image} summarizes the results obtained from $10$-dimensional PCA of the image matrices. 
The MSE values represented in the first two rows of the table show that S-RSVD is more accurate than RSVD. 
To ensure that the results are not due to chance, we run the experiment $30$ times and perform two t-tests with the following null hypotheses:
\begin{itemize}
  \item $H_0^1$:\emph{there is no difference between the MSE of S-RSVD and RSVD.}
  \item $H_0^2$:\emph{there is no difference between the individual column reconstruction errors of S-RSVD and RSVD.}
\end{itemize}
The former hypothesis is validated on the $30$ MSE pairs obtained from the SVD methods, but the later hypothesis is validated on the pairs of the reconstruction error of individual images. 
The $p$-values represented in \TABLE{MSE_image} reject both hypotheses and confirm that the better results obtained from S-RSVD are not by chance. 
The rejection of $H_0^2$ indicates that S-RSVD results in not only lower MSE for the entire image matrices, but also for individual images. 
\begin{table}
  \centering
  \begin{tabular}{lrr|rrrr}
            &\multicolumn{2}{c}{image data} & \multicolumn{4}{c}{word data} \\
            & digits  & faces & $n=\expnum{1}{3}$ & $n=\expnum{1}{4}$ & $n=\expnum{1}{5}$ & $n=\expnum{3}{5}$ \\ \hline
    MSE of S-RSVD  & $\mathbf{415.7}$ & $\mathbf{\expnum{15.3}{7}}$ & $\mathbf{\expnum{195}{-5}}$  & $\mathbf{\expnum{235}{-5}}$  & $\mathbf{\expnum{763}{-5}}$  & $\mathbf{\expnum{994}{-5}}$ \\
    MSE of RSVD    & $430.6$ & $\expnum{16.1}{7}$ & $\expnum{200}{-5}$           & $\expnum{236}{-5}$           & $\expnum{765}{-5}$           & $\expnum{998}{-5}$\\
    $p_1$-value & $0.00$ & $0.00$ & $0.00$  & $0.00$  & $0.00$  & $0.00$\\
    $p_2$-value & $0.00$ & $0.00$ & $0.00$   & $0.00$   & $0.00$   & $0.00$\\
    WR of S-RSVD  & $\mathbf{66}\%$ & $\mathbf{82}\%$  & $\mathbf{71}\%$ & $\mathbf{73}\%$ & $\mathbf{77}\%$ & $\mathbf{70}\%$\\
    WR of RSVD  & $34\%$ & $18\%$ & $29\%$ & $27\%$ & $23\%$ & $30\%$ \\
  \end{tabular}
  \caption{The reconstruction error statistics of image and word data.}
  \label{MSE_image}
\end{table}

\begin{figure}
  \centering
  \subfloat[][]{
    \includegraphics[scale=.5]{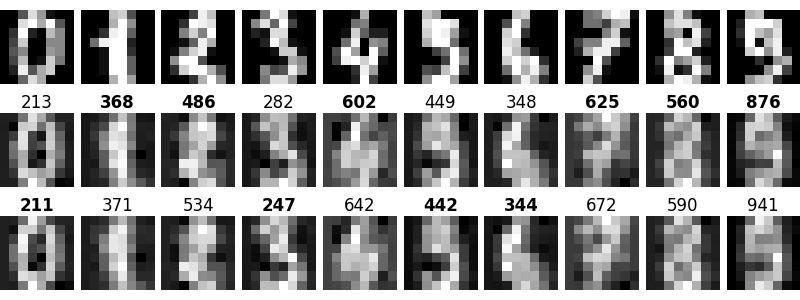}
    \label{fig:digits}
  }

  \subfloat[][]{
    \centering
    \includegraphics[scale=0.5]{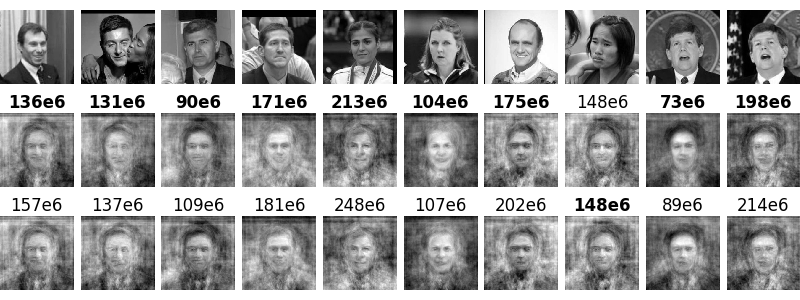}
    \label{fig:digits}
  }
\caption{The effect of mean-centering on (a) handwritten digits and (b) facial data. In each sub-figure, the top row is the original image, the rows in the middle and bottom are the S-RSVD and RSVD reconstructed images, respectively. The reconstruction errors are shown on top of each image.}
  \label{fig:image_data}
\end{figure}

To provide a better picture of how well the SVD algorithms perform on individual images, we plot the first $10$ images of each data matrix and estimate the win-rates of the algorithms. 
\FIG{fig:image_data} shows the examples of the original handwritten and facial images (the top rows), and their reconstructions using S-RSVD (the middle rows) and RSVD (the bottom rows) with the reconstruction error values on top of each image.
For most images, S-RSVD is more accurate than RSVD. 
To generalize this observation, we estimate the win-rate (WR) of the algorithms (\ie the number of images for which one algorithm is more accurate than the other algorithm out of the total number of images).
The results shown in \TABLE{MSE_image} indicate that $66\%$ of the handwritten images and $82\%$ of the facial images are reconstructed more accurately by S-RSVD than RSVD. 

\subsection{Word Data}
In this section, we experiment with word probability co-occurrence matrices whose elements are the probability of seeing a target word in the context of another word.
Our experiments are based on the word co-occurrences probabilities estimated from the English Wikipedia corpus used in the CoNLL-Shared task $2017$.\footnote{\url{https://lindat.mff.cuni.cz/repository/xmlui/handle/11234/1-1989}}
For each target word $w_i$, we estimate the probability of seeing the word conditioned on the occurrence of another word $w_j$, called the context word (\ie $p(w_i|w_j)\approx\frac{n(w_j,w_i)}{n(w_j)}$).
The $i$th column of a probability co-occurrence matrix associated with the word $w_i$ is a distributional representation of the word.

Due to the Zipfian distribution of words and relatively large number of words, a word probability co-occurrence matrix is a large and sparse matrix with a high degree of sparsity. 
A mean subtraction turns the matrix to a dense matrix that needs huge amount of memory and processing time to be analyzed. 
The \textsc{Shifted-Randomized-SVD} algorithm can substantially improve the performance of analyzing a mean-centered co-occurrence matrix. 

We estimate $100$-dimensional PCA representations of different $m\times n$ word probability co-occurrence matrices formed with $m=1000$ most frequent context words and $n$ most frequent target words with different values of $n$. 
Each experiment is run $30$ times with different random seeds. 
The right side of \TABLE{MSE_image} represents the statistics of the reconstruction errors obtained from each of the factorization algorithms. 
The first two rows of the table confirm that S-RSVD is more accurate than RSVD. 
To see whether the difference between MSEs is statistically significant, a t-test with the null hypothesis $H_0^1$:\emph{there is no difference between the MSE of S-RSVD and RSVD} is performed. 
Using the $30$ pairs of MSE values obtained from each of the factorization algorithms, the test rejects the null hypothesis $H_0^1$ with a high confidence level (see $p_1$-value in \TABLE{MSE_image}). 

We study the effect of the mean-centering on the reconstruction of the distributional representation of individual words (\ie each column of the co-occurrence matrix). 
A t-test is performed to validate the null hypothesis $H_0^2$:\emph{there is no difference between the individual column reconstruction errors of S-RSVD and RSVD}.
The acceptance probabilities of $H_0^2$ shown as $p_2$-values in \TABLE{MSE_image} confirm that the differences between the reconstruction errors of individual words is indeed significant.
The win-rates (WR) of each of the algorithms shows that the mean-centering is beneficial to the reconstruction of the majority of words. 

\section*{Conclusion}
We extend the randomized singular value decomposition algorithm of \citet{Halko2011finding} to factorize a shifted data matrix (\ie a data matrix whose columns vectors are shifted by a vector in their eigenspace) without explicitly constructing the matrix in the memory.
With no harm to the performance of the original algorithm on dense matrices, the extended algorithm leads to substantial improvement to the accuracy and efficiency of the algorithm when used for low-rank approximation and principal component analysis of sparse data matrices. 
The algorithm is tested on different types of data matrices including randomly generated data, image data, and word data, with their mean vector as the shifting vector. 
The experimental results show that the extended algorithm results in lower mean squared reconstructions error in all experiments through successfully incorporating the mean-centering step to SVD.

\end{document}